# ALBATROSS: A robotised system for high-throughput electrolyte screening via automated electrolyte formulation, coin-cell fabrication, and electrochemical evaluation

Hyun-Gi Lee[1,‡], Jaekyeong Han[1,‡], Minjun Kwon[1], Hyeonuk Kwon[2],

Jooha Park[3], Hoe Jin Ha[3], Dong-Hwa Seo[1,2,*]

[1]Department of Materials Science and Engineering, Korea Advanced Institute of Science and Technology (KAIST), Daejeon, 34141, Republic of Korea

[2]Graduate School of Green Growth and Sustainability, Korea Advanced Institute of Science and Technology (KAIST), Daejeon, 34141, Republic of Korea

[3]LG Energy Solution, Seoul, 07796, Republic of Korea

[‡]These authors contributed equally: Hyun-Gi Lee and Jaekyeong Han
Corresponding Authors: *e-mail: dseo@kaist.ac.kr (Prof. D.-H. Seo)

## Abstract

As battery technologies advance toward higher stability and energy density, the need for extensive cell-level testing across various component configurations becomes critical. To evaluate performance and understand the operating principles of batteries in laboratory scale, fabrication and evaluation of coin cells are essential processes. However, the conventional coin-cell assembly and testing processes require significant time and labor from researchers, posing challenges to high-throughput screening research. In this study, we introduce an Automated Li-ion BAttery Testing RObot SyStem (ALBATROSS), an automated system capable of electrolyte formulation, coin-cell assembly, and electrochemical evaluation. The system, integrated within a argon-filled glovebox, enables fully automated assembly and testing of up to 48 cells without researcher intervention. By incorporating custom-designed robot gripper and 3D-printed structures optimized for precise cell handling, ALBATROSS achieved high assembly reliability, yielding a relative standard deviation (RSD) of less than 1.2% in discharge capacity and a standard deviation of less than 3 Ω in EIS measurements for NCM811∥Li half cells. Owing to its high reliability and automation capability, ALBATROSS allows for the acquisition of high-quality coin-cell datasets, which are expected to accelerate the development of next-generation electrolytes.

**Keywords:** Laboratory Automation, Autonomous Laboratory, Self-Driving Laboratory, AI-Robotics, coin-cell assembly, liquid electrolyte screening, high-throughput screening

# Introduction

Liquid electrolytes are one of the key components that govern the operation of lithium-ion batteries (LIBs). The liquid electrolytes serve as the medium for ion transport and govern high-rate capability, energy efficiency, the formation of interfacial layers such as the solid electrolyte interphase (SEI), and the thermal and chemical stability of the cell. To improve these properties, various types of electrolytes have been developed, and a wide range of salts, solvents, and additives have been created and utilized. It is anticipated that future electrolyte development will require careful selection and optimal combination of various salts, solvents, and additives[1,2]. Therefore, an efficient method for evaluating diverse electrolytes is needed, and this is typically done by assembling and testing coin cells, one of the smallest types of battery. However, assembling and testing a large number of coin cells requires significant time and labor from researchers, posing a challenge.

To address the need for high-throughput cell fabrication, several automated coin-cell assembly systems for LIBs with non-aqueous liquid electrolytes utilizing robotics within argon-filled glovebox have been reported, including AutoBass[3], Poseidon[4], and ODACell[5]. These systems are designed to assemble numerous cells using pre-loaded cell components with minimal human intervention. Notably, both AutoBass and Poseidon are limited to the assembly process; subsequent electrochemical evaluations such as cycle testing must be performed manually by the researcher. ODACell extends this functionality by automating both cell assembly and cycling test. However, the system is equipped with 16 cycling channels, allowing the cycling of 16 cells at a time. Because cycling tests are time-consuming, acquiring a large dataset requires considerable time. Therefore, to solve this bottleneck, the automated cycling modules must be expanded to enable more efficient data acquisition.

On the other hand, while charge-discharge curves provide essential information about the overall performance of a cell, they offer limited insight into the underlying degradation mechanisms. Electrochemical Impedance Spectroscopy (EIS), which enables the interpretation of bulk resistance, charge transfer resistance, and SEI resistance, can serve as a complementary analytical tool for analyzing cell degradation[6,7]. Despite its diagnostic utility, obtaining EIS data for multiple cells is often limited because frequent manual intervention is required to conduct both cycling and impedance measuring steps. As a result, EIS is usually performed only on a small number of selected cells, either immediately after formation or after long-term cycling for lifetime assessment. To overcome these constraints, the development of an integrated system capable of automatically performing both charge-discharge cycling test and EIS analysis is important. Such a system would enable the acquisition of comprehensive electrochemical and impedance data across a large number of cells with minimal manual effort, thereby facilitating a deeper understanding of the underlying principles of cell degradation.

The aforementioned challenges can be addressed through the following key strategies:

1. Systematic formulation of electrolytes with various combinations and ratios of salts, solvents, and additives.

2. Automated assembly of a large number of cells without human intervention.

3. Full automation of charge-discharge testing and EIS measurements for the assembled cells.

In this work, to implement these strategies, we developed Automated Li-ion BAttery Testing RObot SyStem (ALBATROSS), an integrated, fully automated platform capable of assembling and evaluating up to 48 coin cells without researcher involvement (Fig. 1). The ALBATROSS performs electrolyte formulation, coin-cell assembly, charge-discharge cycling, and EIS measurements within an argon-filled glovebox environment. The platform incorporates 48 cycling channels and 2 EIS channels, enabling the efficient collection of both electrochemical and impedance data. The system is capable of formulating an electrolyte, assembling a coin cell, and initiating a cycling test within 4 minutes, and it successfully assembled 85 out of 87 cells during validation. The system also demonstrated high assembly repeatability, yielding a relative standard deviation (RSD) of less than 1.2% in discharge capacity and a standard deviation below 3 $\Omega$ in EIS measurements. To achieve this level of automation, a variety of custom-designed robot gripper and 3D-printed structures were developed and integrated into the platform. With ALBATROSS, extensive cycling and impedance datasets can be obtained across diverse electrolyte composition spaces, providing insights into the fundamental roles of electrolytes and accelerating electrolyte development. Moreover, this platform is expected to extend beyond liquid electrolyte screening to address a broader range of challenges, including the optimization of formation protocols for SEI design[8], and the exploration of fast charge-discharge protocols[9].

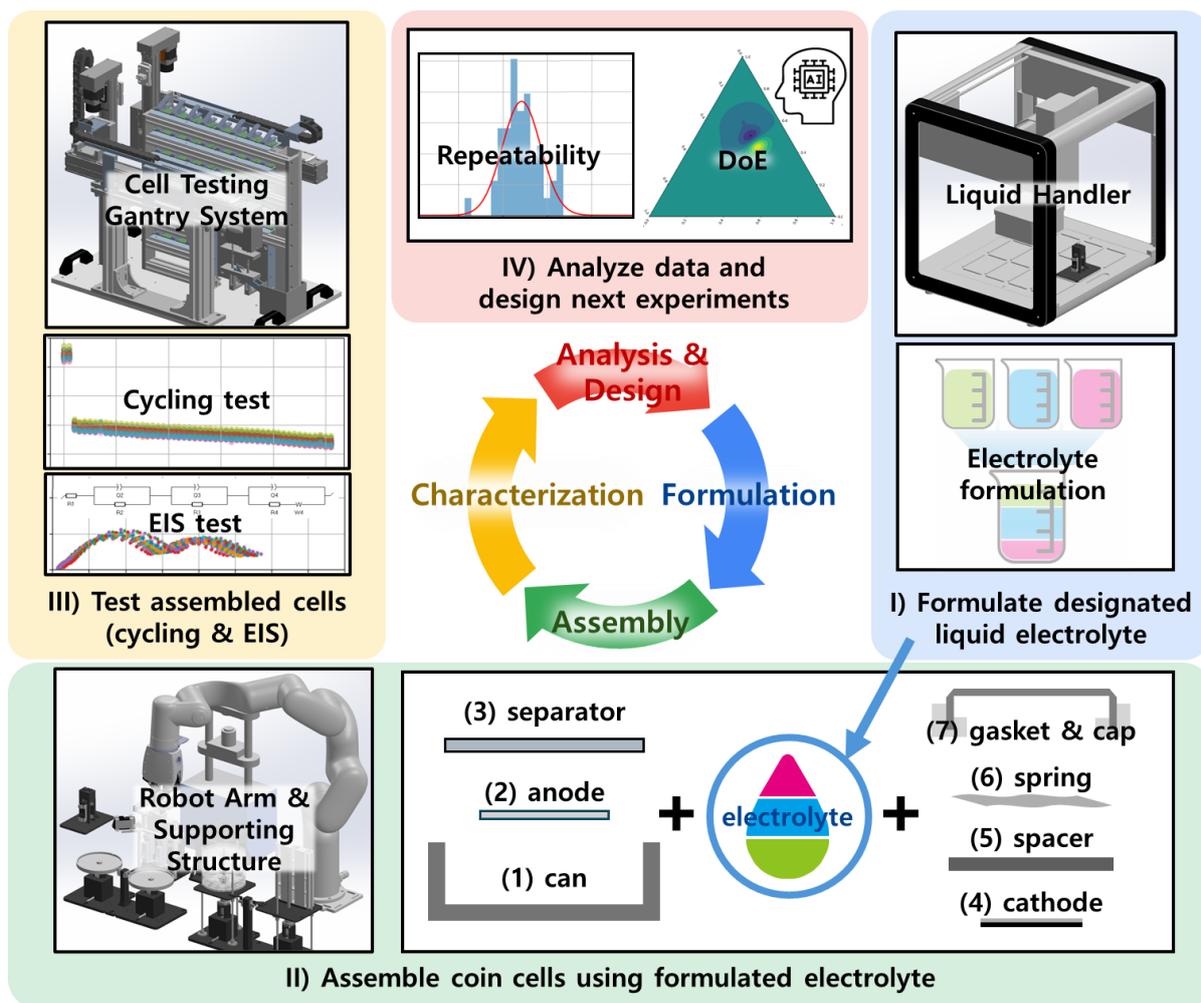

**Figure 1. Illustration of ALBATROSS flow:** I) Formulating liquid electrolyte, II) Assembling coin cells using formulated electrolyte, III) Testing assembled cells, IV) analyzing data and designing the next experiments.

## Methods

**Devices and Materials**

The automated system for electrolyte formulation, cell assembly, and evaluation was constructed using a 6-axis robot arm (xArm6, UFactory), a liquid handler (OT-2, Opentrons), an automated coin cell crimper (MSK-160E, MTI), customized with an elevated vertical axis to ensure unobstructed access by the robot arm. Additional actuating and motion components including stepper motors (C-42STM03, Misumi) and linear actuators (12LF-12F-27 and 12LF-17F-40, Mightyzap) were utilized. System control and synchronization were managed using a programmable logic controller (PLC; NX102-9000, Omron). Custom structural parts were designed in CAD and fabricated using a 3D printer (A15CR, Cubicon) with PLA filament or from aluminum using CNC machining when necessary. Cycling test equipment (Neware, CT-4008T-5V50mA-164) and impedance measurement devices (Biologics, SP-150e) were installed inside the glovebox.

For the reproducibility tests, 2032-type coin cell cases (MTI) were used as cell enclosures. The cathode consisted of a 12.5 mm diameter NCM811 electrode (MTI), and the anode was a 16 mm diameter, 0.1 mm thick lithium metal foil (Honjo). A 19 mm diameter glass fiber separator (Whatman) was employed, and each cell was filled with 70 μL of electrolyte composed of 1 M $LiPF_6$ in EC:EMC (3:7 vol%) with 2 wt% VC additives (EnChem). All components were handled within an argon-filled glovebox to prevent moisture or oxygen contamination.

**Electrolyte formulation**

In this system, electrolyte formulation is performed using the liquid handler. To prepare electrolytes with various compositions, concentrated salt solutions were diluted with pure solvents to the desired concentrations. For solvents such as ethylene carbonate (EC), which are solid at room temperature, a temperature-control module inside the liquid handler is used to maintain the solvent at 60°C prior to dispensing. During aspiration, any excess solution adhering to the pipette tip is removed by touching the tip to the vial entrance. The solutions are then mixed using the automated micropipette, which performs 20 repetitive mixing strokes over a total duration of 3 minutes.

**Electrochemical analysis**

To evaluate the reproducibility of cell cycling, two formation cycles followed by 50 main cycles were conducted for all assembled cells. The formation cycles were conducted within a voltage range of 3.0–4.3 V at a 0.1 C-rate, using a constant current–constant voltage (CCCV) charging and a constant current (CC) discharging mode. Subsequently, the main cycles were performed within a range of 3.0–4.2 V at a 1 C-rate using CC charging and discharging mode. For comparative analysis, 40 cells manually assembled by a researcher and 40 cells assembled using the ALBATROSS were tested under identical conditions.

To verify the reproducibility of EIS measurements, an additional set of 45 newly assembled cells was evaluated using the automated system. Each cell underwent two formation cycles at 0.1C between 3.0–4.3 V, followed by two charge-discharge cycles at 0.5, 1, 2, and 3 C-rates within the 3.0–4.2 V range. EIS measurements were performed after completing two cycles at each C-rate condition, following a 30-minute rest period, over a frequency range of 200 kHz to 0.1 Hz. If two cells enter the EIS-measurement waiting status simultaneously, the second cell initiates its EIS measurement immediately after the first cell completes its measurement.

## Results and Discussion

ALBATROSS integrates a liquid handler, a robot arm, potentiostats, and EIS modules within a four-port glovebox, enabling fully automated electrolyte formulation, coin-cell assembly, and electrochemical evaluation without human intervention. The system assembled coin cells with a success rate of 97.7% (85 out of 87 cells). A single operation, which included electrolyte formulation, cell assembly, and initiation of the cycling protocol, required approximately 4 minutes, allowing the system to process 48 cells in roughly 200 minutes. Assuming that the electrochemical cycling procedure requires approximately six days, the platform is capable of generating data for up to 240 cells per month.

In addition to the capability for high-throughput data generation, a critical aspect of any automated system is the achievement of high reproducibility. High reproducibility serves as an indicator that, by leveraging the precise and consistent operation of mechanical components, experimental errors and variations that may arise from manual handling by researchers can be minimized, thereby enabling the acquisition of high-quality data. Coin-cell assembly and evaluation involve the handling of diverse cell components and the management of charge-discharge processes for multiple cells, making considerable effort necessary to ensure reproducibility. In the following sections, we describe the strategies implemented to secure high-quality data and present the corresponding test results.

The developed ALBATROSS is composed of three main automated modules: (I) electrolyte formulation, (II) coin cell assembly using the formulated electrolyte, and (III) electrochemical evaluation of the assembled cells (Figs. 2, 3). The electrolyte formulation module utilizes a liquid handler to enable automated mixing and dispensing of electrolyte components. The cell assembly module is operated by a robot arm, while the electrochemical testing module is constructed based on gantry systems integrated with potentiostat and impedance measurement devices. Unlike conventional glovebox-integrated automation systems, which often employ compact 3-axis robots due to limited internal space, the ALBATROSS adopts a 6-axis robot arm with a wide working range (700mm). This design choice was made to accommodate the simultaneous assembly of a large number of cells and the handling of various devices and associated waste, which requires broader operational range. However, employing a 6-axis robot arm within the confined space of a glovebox makes challenges related to its limited maneuverability. To mitigate this, the system incorporates two key strategies: (1) the development of a custom gripper capable of handling multiple types of components with a simple mechanical structure, and (2) the implementation of support structures that spatially arrange the cell components within the robot's effective working area. A detailed discussion follows in subsequent sections.

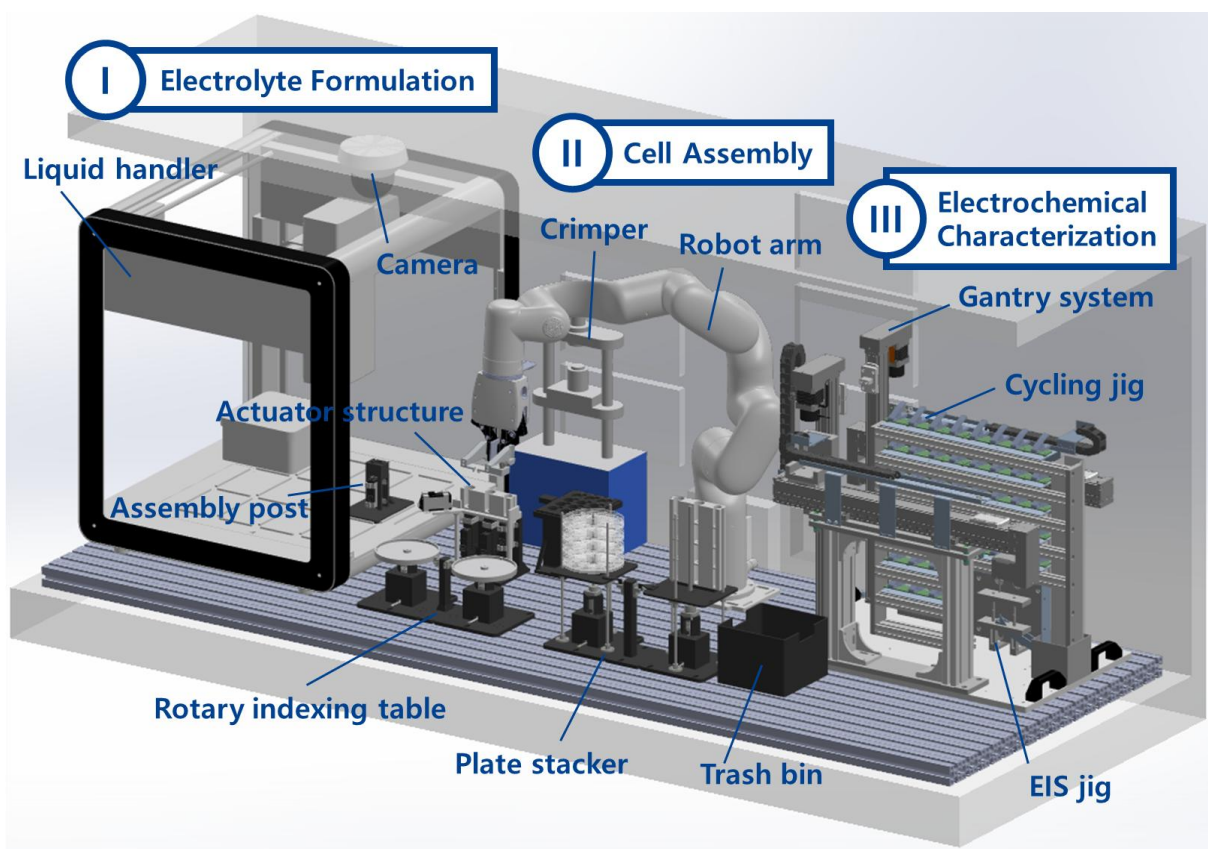

**Figure 2. Schematic illustration of ALBATROSS setup.** I) Electrolyte formulation and dispense part using the liquid handler. II) Coin-cell assembly part using the robot arm. III) Coin-cell electrochemical characterization part using two gantry systems, potentiostats and impedance measurement instruments.

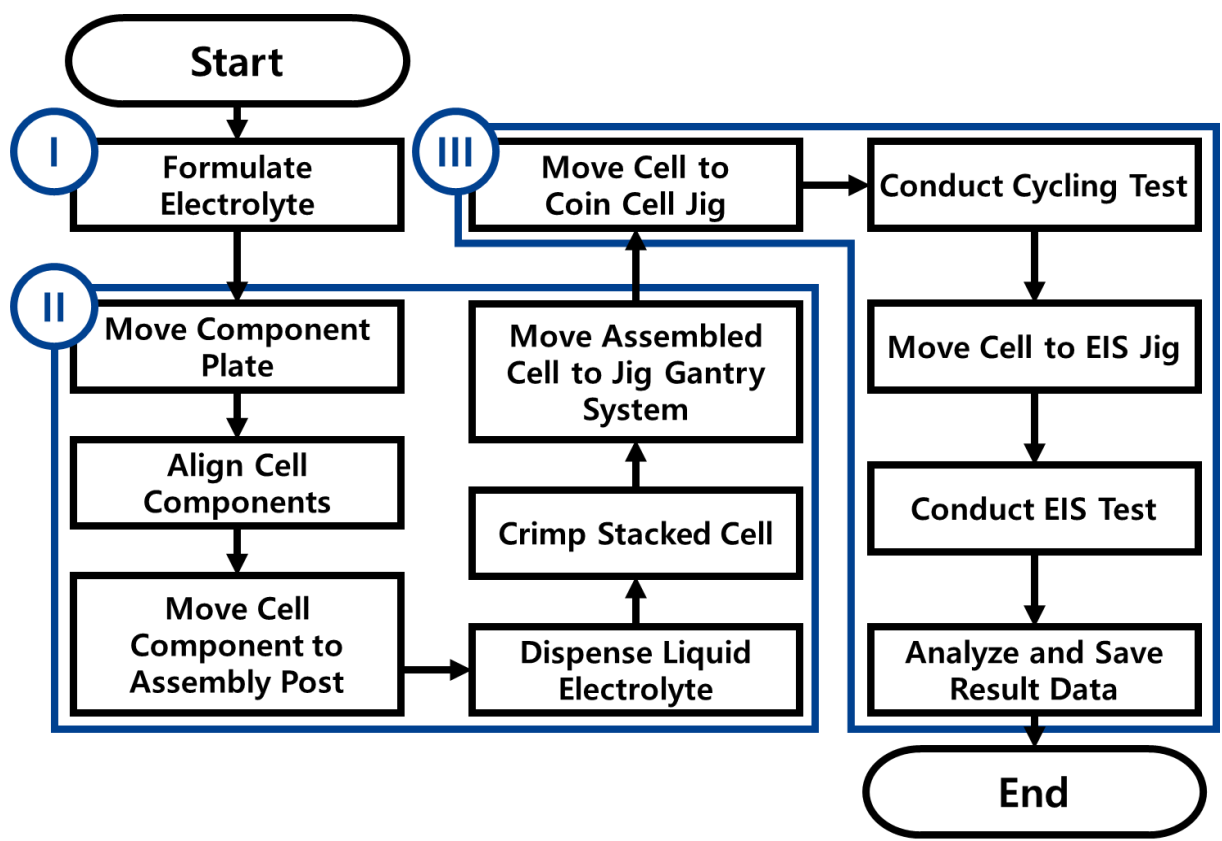

**Figure 3. Flowchart of ALBATROSS.** I) Electrolyte formulation II) Coin-cell assembly III) Coin-cell electrochemical characterization.

**Coin-cell assembly**

The coin-cell assembly and evaluation processes were conducted using the robot arm. To enable the automated assembly and evaluation of 48 cells simultaneously, the robot arm was designed with a gripper capable of accommodating both a parallel gripper and a vacuum gripper (Fig. S4). The vacuum gripper was used to transfer the coin cell components, such as the can, anode, and separator, after which the liquid dispenser precisely dispensed the electrolyte into the cell. Subsequently, the cathode, spring, spacer, and cap were transferred and then assembled using an automated crimper, completing the coin-cell assembly. Since transferring the spring was difficult with the vacuum gripper, the parallel gripper was utilized for this task.

The system was installed within a glovebox, making robot-coordinate calibration challenging and necessitating simplification of the calibration process. Simply arranging the components at fixed equal distances would make future recalibration difficult, particularly considering that there are 48 sets of 7 different components, totaling 336 components. Such an arrangement could not only complicate coordinate recalibration but also impede the robot arm's accessibility. To facilitate reliable handling of various components and simplify coordinate calibration, supporting structures were fabricated to supply all components to the robot arm at fixed positions (Figs. 2, S1, S2). Using rotary and stackable structures, cell components pre-arranged in sets of 12 were consistently positioned to enhance robot accessibility and repeatability. This configuration ensured that components were always transferred to the same location, which is critical as precise component placement strongly influences subsequent cell performance. It also reduced the number of robot's calibration points from 336 to 8, dramatically improving recalibration efficiency, which are critical issues for systems operating inside a glovebox. The use of these supporting structures substantially improved the reproducibility of the assembled cells.

The system assembles 48 cells autonomously by utilizing four component plates, each containing 12 parts. Positioning these plates for robotic handling posed challenges because plates occupy a relatively considerable volume, require unobstructed access for robotic gripping, and must be handled within the confined space of a glovebox, where maneuverability is inherently limited. As a result, the plate-storage structure, referred to here as the plate stacker, had to fulfill two competing requirements: occupying minimal space while simultaneously elevating the plates to a height that enables easy robotic access and providing sufficient mechanical support to prevent plate tilting or collapse. As is common in specialized automation systems, no commercially available structure could satisfy these constraints. Therefore, we disassembled a cartesian robot and repurposed the robot components to design a custom storage mechanism. The resulting structures consistently positioned each plate at an elevated and robot-accessible location (Figs. 2, S1, S3). This customized solution effectively mitigated the accessibility limitations imposed by the glovebox environment and reduced the likelihood of operational failures due to spatial interference or unstable component positioning. Consequently, the plate stacker enhances the overall operational reliability of the automated assembly workflow.

**Cell Characterization**

After cell assembly process, the assembled cells were then transported from the crimper to the cycling gantry system, which consisted of two gantries designed for automated cycling and EIS evaluation (Figs. 2, S6, S7). The automated cycling and EIS system was configured with 48 potentiostat channels and 2 EIS channels. Directly transferring the coin cells from the robot arm to all 48 potentiostat channels or 2 EIS channels is challenging, as it requires simultaneous control of the jig opening and cell placement. To address this, two gantry systems were employed to receive the coin cells from the robot arm at the fixed position and reliably transfer them to vacant cycling channels. This process operates independently of the robot arm, substantially enhancing the efficiency of the workflow from assembly to evaluation.

As a result, the entire process, from electrolyte formulation to electrochemical evaluation, was fully automated without human intervention. Moreover, the system enabled EIS measurements to be automatically performed during cycling, not only for a subset of cells but for all 48 cells, allowing detailed monitoring of degradation behavior and the acquisition of high-quality experimental data.

**Reproducibility test**

To evaluate the performance of the completed system, 40 cells were assembled using the ALBATROSS and another 40 cells were manually assembled by a researcher under identical electrolyte compositions. All cells underwent cycling tests with the same protocol. For reproducibility analysis, the average, standard deviation, and relative standard deviation (RSD) of the formation cycle and 50th of the main cycles data were calculated. The RSD serves as an indicator of system reproducibility, with values below 1% considered to represent excellent consistency.

$$\bar{x} = \frac{\sum x}{n}, \qquad s = \sqrt{\frac{\sum(x - \bar{x})}{n - 1}}, \qquad \widehat{c_v} = \frac{s}{\bar{x}}$$

($\bar{x}$: mean of sample, $n$: the number of sample, $s$: sample standard deviation, $\widehat{c_v}$: coefficient of variance (CV) or relative standard deviation (RSD))

The 40 cells manually assembled by the researcher exhibited a RSD of 0.955% in the formation cycle and 1.142% in the 50th cycle. Similarly, the 40 cells assembled using the ALBATROSS showed a RSD of 1.040% in the formation cycle and 1.210% in the 50th cycle, which are comparable to the manually assembled results. (Fig. 4) These results demonstrate that the ALBATROSS achieved a level of reproducibility comparable to that of an experienced researcher, confirming the reliability of the automated assembly process.

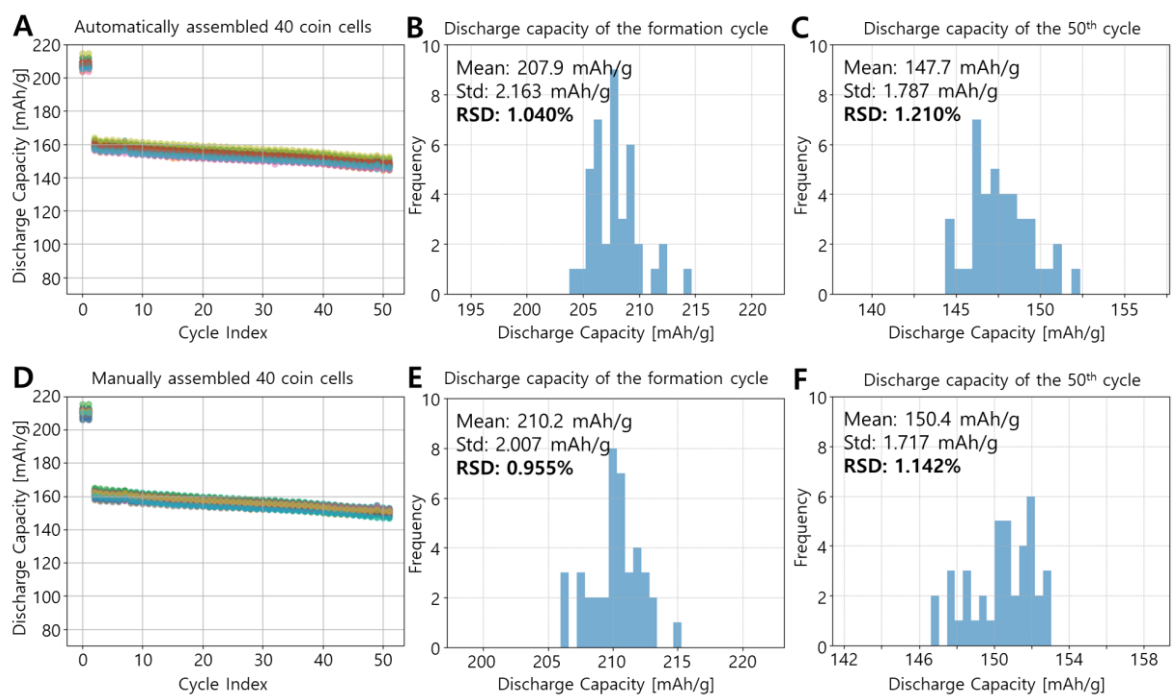

**Figure 4. Discharge-capacity profiles and reproducibility results for the formation and the 50th cycle of 40 cells assembled by ALBATROSS and by a human researcher.** (A–C) Cycle test results of 40 cells assembled by ALBATROSS and (D–F) results of cells assembled manually. (A, D) Discharge-capacity profiles of 40 cells, (B, E) histogram of discharge capacity of the formation cycle, and (C, F) histogram for the 50th cycle.

**Automated impedance measurement**

For EIS evaluation, 45 newly assembled cells were tested under 0.5C, 1C, 2C, and 3C cycling conditions. The capacity reproducibility exhibited similar trends to those observed in the previous tests (Fig. 5). Subsequent EIS measurements were then conducted for the same cells. The EIS data were analyzed based on an equivalent circuit model composed of three RC elements, representing the electrolyte resistance ($R_1$, $R_{bulk}$), the contact resistance between active materials and current collectors ($R_2$, $R_C$), the SEI resistance ($R_3$, $R_{SEI}$), and the charge transfer resistance ($R_4$, $R_{CT}$)[10] (Fig. 6A). In contrast to the cycling capacity results, the EIS data showed relatively large variations among cells (Figs. 6, S9–S12). These findings indicate that even cells exhibiting small deviations in capacity performance can show substantial variation in the RSD of impedance data. This also suggests that impedance measurements collected after cycling can provide meaningful insight into the reasons for capacity loss associated with cell degradation. To the best of our knowledge, reproducibility data for EIS measurements across multiple cells have not been reported previously. Therefore, the present results are expected to provide a valuable reference for understanding the background noise inherent in EIS analysis.

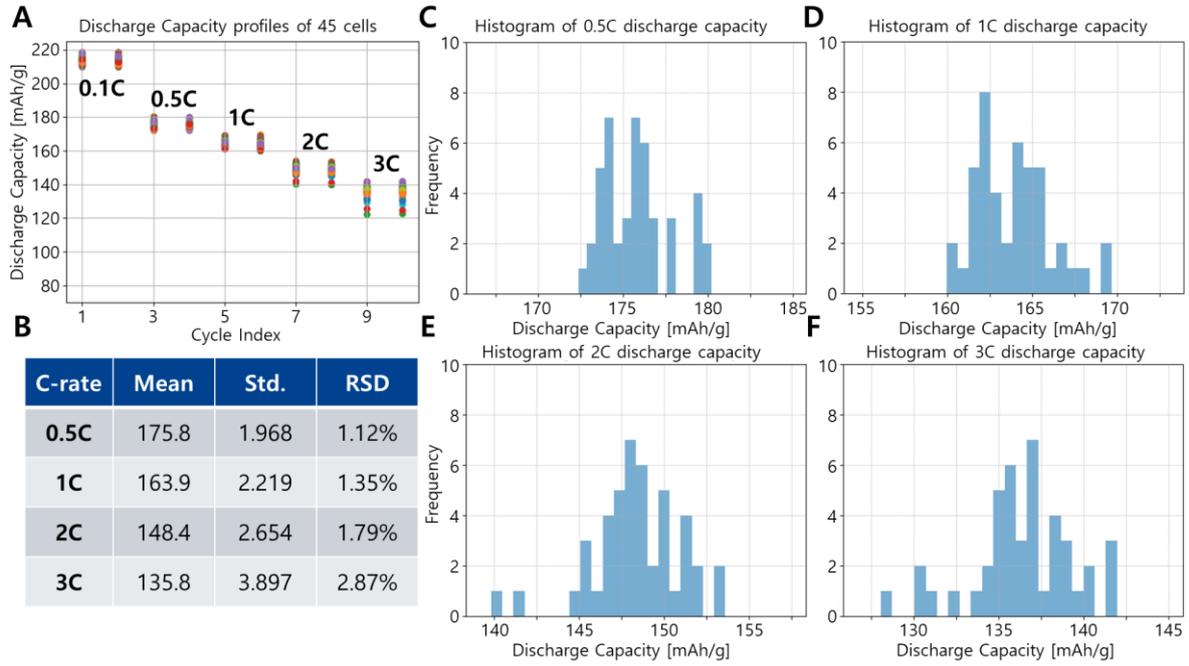

**Figure 5. Cycling results of 45 cells used for EIS evaluation.** (A, B) Discharge capacity of 45 cells cycled at 0.5C, 1C, 2C, and 3C. (C–F) Histograms of discharge capacity distributions at each rate.

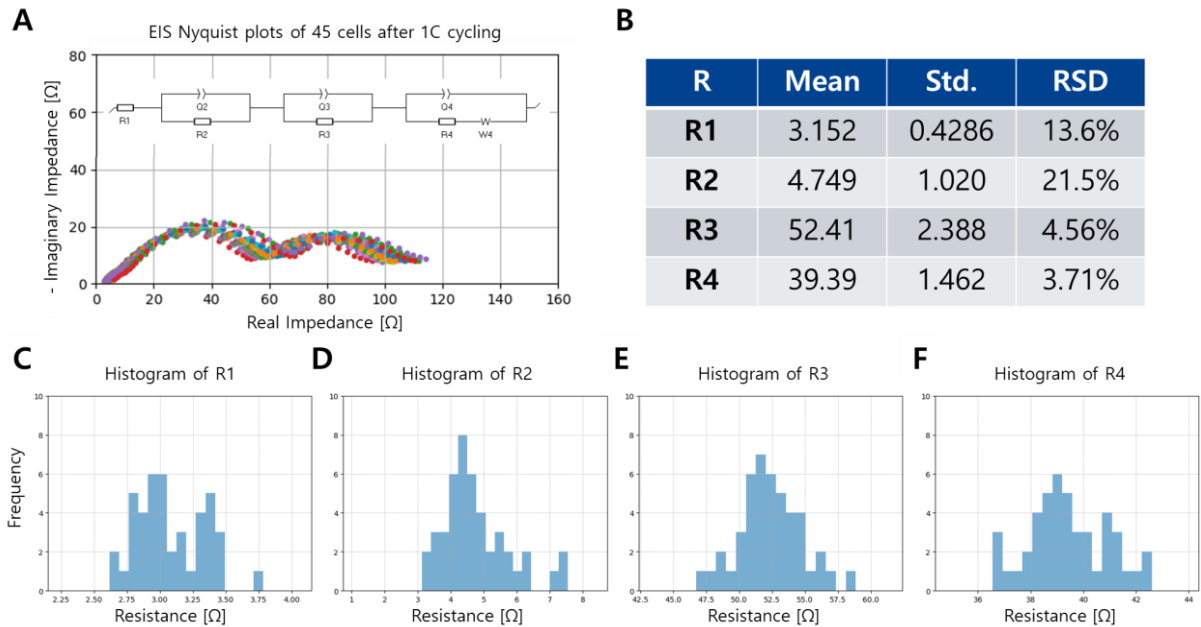

**Figure 6. EIS results of 45 cells after 1C-rate cycling.** (A) Schematic of the equivalent circuit and Nyquist plots for 45 cells and (B) mean, standard deviation, and RSD results of fitted resistance values. (C–F)

Histograms of resistance distribution for different resistance components (R1, R2, R3, and R4).

**Future direction for high-throughput electrolyte discovery**

Despite the development of the ALBATROSS, the design space for electrolytes remains extremely large. In fact, the primary bottleneck in electrolyte evaluation is not the fabrication of the electrolyte or cell assembly, but the long-term evaluation of cell. To efficiently identify promising electrolytes, three key strategies are required: (1) defining an appropriate variable domain, (2) conducting design of experiments through artificial intelligence models, and (3) developing accelerated charge-discharge protocols for rapid cell evaluation. Accordingly, a feasible approach involves selecting candidate composition domains through first-principles calculations, followed by rapid exploration using AI-driven experimental design models based on experimental data obtained from the automated system. Furthermore, developing reliable AI models capable of predicting long-term cycle life from short-term charge-discharge tests[11] and EIS measurements[7] is essential.

# Conclusion

In this study, we introduced ALBATROSS, an automated platform for coin-cell assembly and electrochemical evaluation. The system integrates three major modules: electrolyte formulation and dispensing, coin-cell assembly, and electrochemical characterization. This configuration enables fully automated cell fabrication and evaluation of up to 48 cells without manual intervention. To ensure reliable operation, custom-designed grippers and support structures were implemented to align and handle cell components, resulting in highly reproducible results in cell capacities. In addition, the system can perform EIS measurements on all 48 cells during cycling, providing comprehensive impedance data that could reveal the underlying mechanisms of cell degradation. We anticipate that the high-quality data generated by ALBATROSS will accelerate the discovery of electrolytes, particularly for challenging battery systems such as lithium-metal, lithium-sulfur, and sodium-ion systems.

# Data availability

All the codes, program files and videos for ALBATROSS can be found at https://github.com/Hyun-Gi/ALBATROSS.

## Author contributions

Hyun-Gi Lee contributed to modeling, hardware fabrication, software development, literature review, manuscript writing, manuscript review and revision, and data visualization. Jaekyeong Han contributed to hardware modification, software development, literature review, and manuscript review and revision. Minjun Kwon contributed to hardware fabrication and software development. Hyeonuk Kwon contributed to manual cell data generation, literature review, and software modification. Jooha Park contributed to literature review and manuscript review and revision. Hoe Jin Ha contributed to literature review and manuscript review and revision. Dong-Hwa Seo contributed to methodology and conceptualization, idea development, manuscript review and revision, and overall supervision of the project.

## Conflicts of interest

There are no conflicts to declare.

## Acknowledgements

This research was supported by the LG Energy Solution.